\documentclass[
]{ceurart}

\usepackage[norelsize]{algorithm2e}
\usepackage{algorithmic}
\begin{document}

\copyrightyear{2021}
\copyrightclause{Copyright for this paper by its authors.
  Use permitted under Creative Commons License Attribution 4.0
  International (CC BY 4.0).}

\conference{Forum for Information Retrieval Evaluation, December 15-18 2023, India}

\title{Prompted Zero-Shot Multi-label Classification of Factual Incorrectness in Machine-Generated Summaries}

\author[1]{Aniket Deroy}[%
%orcid=0000-0002-0877-7063,
email=roydanik18@kgpian.iitkgp.ac.in,
%url=https://yamadharma.github.io/,
]
\author[1]{Subhankar Maity}[%
%orcid=0000-0002-0877-7063,
email=subhankar.ai@kgpian.iitkgp.ac.in,
%url=https://yamadharma.github.io/,
]
\author[1]{Saptarshi Ghosh}[%
%orcid=0000-0002-0877-7063,
email=saptarshi.ghosh@gmail.com,
%url=https://yamadharma.github.io/,
]
%\cormark[1]
\address[1]{IIT Kharagpur, Kharagpur, India}
\begin{abstract}
  This study addresses the critical issue of factual inaccuracies in machine-generated text summaries, an increasingly prevalent issue in information dissemination. Recognizing the potential of such errors to compromise information reliability, we investigate the nature of factual inconsistencies across machine-summarized content. We introduce a prompt-based classification system that categorizes errors into four distinct types: misrepresentation, inaccurate quantities or measurements, false attribution, and fabrication. The participants are tasked with evaluating a corpus of machine-generated summaries against their original articles. Our methodology employs qualitative judgements to identify the occurrence of factual distortions. The results show that our prompt-based approaches are able to detect the type of errors in the summaries to some extent, although there is scope for improvement in our classification systems.
\end{abstract}


\maketitle

\section{Introduction}

In an era where information dissemination is predominantly driven by digital platforms, the accuracy and integrity of content have become paramount. Machine-generated summaries, designed to distill complex articles into digestible formats, have gained traction because of their efficiency and scalability. However, the susceptibility of these systems to introduce factual errors poses a significant challenge. This research endeavors to meticulously analyze the prevalence of factual inaccuracies within machine-generated summaries by establishing a systematic methodology for identification and categorization.

We propose a novel prompt-based framework~\cite{ding2022openprompt} that empowers participants to discern and classify factual inaccuracies into one of four distinct types: misrepresentation, inaccurate quantities or measurements, false attribution, and fabrication. Each category embodies unique characteristics of factual errors, ranging from subtle misinterpretations to the deliberate creation of non-existent facts. Misrepresentation refers to the skewed presentation of information that can alter the perceived meaning. Inaccuracies in quantities or measurements involve numerical or statistical deviations from the truth. False attribution represents the erroneous association of statements or actions with individuals or entities. Lastly, fabrication denotes the most egregious breach, where information is concocted without any factual foundation.

This research serves as a critical investigation into the fidelity~\cite{zhangevaluating} of machine-generated summaries. By scrutinizing these summaries against their source articles, we aim to quantify the extent of factual distortions and understand their implications. The ultimate goal is to enhance the credibility of machine-generated content, ensuring that it serves as a reliable conduit for knowledge and information in the digital age. Our results show that our novel prompt-based approaches are capable of detecting the type of errors in the summaries to some extent, although there is scope for improvement for our classification systems.
The task is based on ~\cite{ACM_ILSUM23,ILSUM_overview} which are the original track papers.
\section{Related Work}

In recent years, the field of natural language processing (NLP) has seen a significant shift towards the development and utilization of large language models (LLMs). These LLMs, particularly exemplified by OpenAI's GPT series (Generative Pre-trained Transformer), have revolutionized various NLP tasks. The foundational concept behind these models involves pre-training on vast amounts of text data, enabling them to learn intricate language patterns and structures.

Zero-shot prompting with LLMs has been leveraged across various tasks. In text generation, these LLMs exhibit the capacity to produce coherent and contextually relevant content even when prompted by unseen topics or styles. Translation tasks \cite{r1} benefit from zero-shot capabilities, allowing language conversion without specific paired training data. Sentiment analysis \cite{r2}, intent classification \cite{r3}, named entity recognition \cite{r4,r5}, and multi-label text classification \cite{r8} are among other tasks where LLMs prompted in a zero-shot manner showcase robust performance without explicit task-oriented training. Furthermore, question-answering \cite{r7} and summarization tasks \cite{r6} witness effective output through the zero-shot prompting approach, offering pertinent answers and concise summaries without task-specific fine-tuning.

GPT-3.5 Turbo represents a significant advancement in the landscape of LLMs. It builds on the foundation laid by GPT-3 \cite{NEURIPS2020_1457c0d6}, showcasing scale and potential improvement in training methodologies, although specific details regarding the “Turbo” improvements remain proprietary to OpenAI. GPT-3.5 Turbo's training involves self-supervised learning on an extensive and diverse corpus of Internet text, refining its language understanding and generation capabilities. Leveraging the zero-shot learning paradigm, it excels in performing various natural language processing tasks without specific fine-tuning, a hallmark feature carried forward from the GPT-3 architecture. In the context of detecting factual incorrectness in machine-generated summaries, the zero-shot prompting method utilizing LLMs presents a promising approach. The methodology involves instructing the model with label descriptions and tasks, allowing it to identify and classify factual inaccuracies without direct training on specific datasets. GPT 3.5 Turbo, known for its advanced zero-shot learning capabilities, stands as a potential solution for discerning factual errors in machine-generated content.

%\begin{itemize}
%\item \verb|natbib.sty| for citation processing;
%\item \verb|geometry.sty| for margin settings;
%\item \verb|graphicx.sty| for graphics inclusion;
%\item \verb|hyperref.sty| optional package if hyperlinking is required in
%  the document;
%\item \verb|fontawesome5.sty| optional package for bells and whistles.
%\end{itemize}

%All the above packages are part of any
%standard \LaTeX{} installation.
%Therefore, the users need not be
%bothered about downloading any extra packages.
\section{Dataset}
We have been provided with the original articles and incorrect summaries in the training set and the testing set of ILSUM task 2. There are 8497 articles in the train set and 200 articles in the test set.
\section{Task Definition}
The task focuses on identifying factual errors in machine-generated summaries. The objective is to categorize each datapoint into different categories based on factual incorrectness in the summaries.

Possible types of factual incorrectness:

\begin{itemize}

\item \textbf{\textit{Misrepresentation:}}
This involves presenting information in a way that is misleading or gives a false impression. This could be done by exaggerating certain aspects, understating others, or twisting facts to fit a particular narrative.

\item  \textbf{\textit{Inaccurate Quantities} or \textit{Measurements}:} Factual incorrectness can occur when precise quantities, measurements, or statistics are misrepresented, whether by error or intent.

\item  \textbf{\textit{False Attribution}:} Incorrectly attributing a statement, idea, or action to a person or group is another form of factual incorrectness.

\item  \textbf{\textit{Fabrication}:} Making up data, sources, or events is a severe form of factual incorrectness. This involves creating “facts” that have no basis in reality.

\end{itemize}

\section{Methodology}
\subsection{Why Prompting?}
Prompting is a valuable approach to solving multilabel classification problems for several reasons:
\begin{itemize}

\item[--] \textbf{Natural Language Bridge:}
Prompting allows the use of natural language to bridge the gap between machine learning models and complex tasks without the need for extensive reprogramming or model redesign. It essentially converts the classification task into a text generation problem, which large language models are inherently good at solving.

\item[--] \textbf{Transfer Learning:}
Through prompting, models that have been trained on vast datasets can apply their learned knowledge to classify data across multiple labels. This transfer learning is efficient because it leverages pre-existing knowledge without the need for extensive additional training on specialized datasets.

\item[--] \textbf{Flexibility:}
Prompt-based approaches are highly flexible and easily adapted to different tasks and domains. This is particularly useful in multilabel classification, where the relationships and distinctions between categories can be nuanced and context-dependent.

\item[--] \textbf{Efficiency:}
Prompting can reduce the need for large annotated datasets that are typically required to train multi-label classifiers. By using prompts, models can often make predictions without any examples(Zero-Shot classification).

\end{itemize}
\subsection{Prompting approach}

\begin{figure}[h!]
  \centering
  \includegraphics[width=0.70\linewidth]{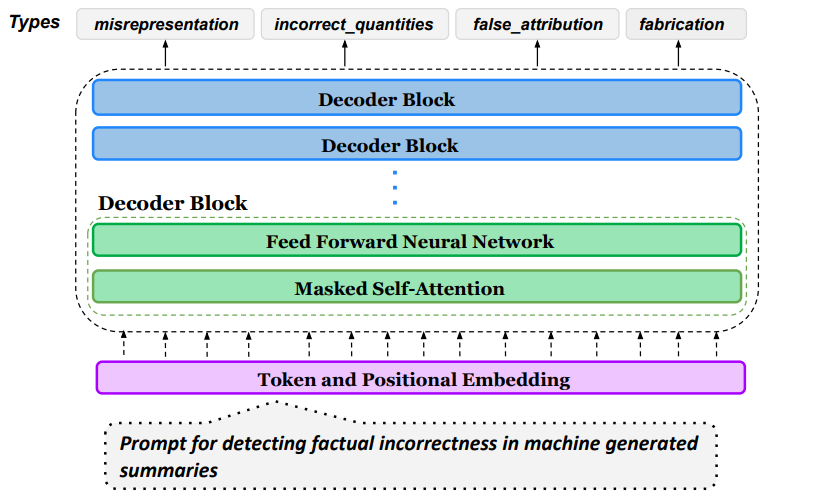}
  \caption{An overview of GPT for zero-shot multi-label classification of factual incorrectness in machine-generated summaries.} \label{fig1}
  %\Description{A woman and a girl in white dresses sit in an open car.}
\end{figure}

\begin{figure}[h!]
  \centering
  \includegraphics[width=0.80\linewidth]{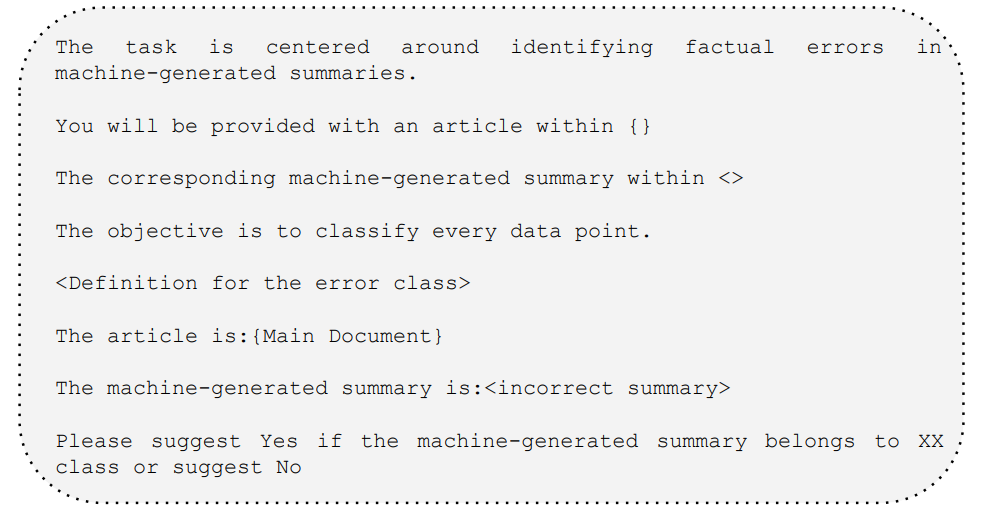}
  \caption{Prompt used for GPT-3.5 Turbo. Where, XX can be misrepresentation, fabrication, false\_attribution, and incorrect\_quantities.} \label{fig2}
  %\Description{A woman and a girl in white dresses sit in an open car.}
\end{figure}

The prompting approach involved employing the GPT-3.5 Turbo model in zero-shot mode for the multi-label classification task of detecting factual incorrectness in machine-generated summaries. The approach included instructing the GPT-3.5 Turbo model in zero-shot mode, providing a set of label descriptions, and outlining the task to be executed. The hyperparameters are as follows: \texttt{temperature} = \{0.5, 0.6, 0.7, 0.8, 0.9\}, \texttt{max-tokens} = 50, and \texttt{stop} = None. A diagrammatic representation of the model is shown in Figure~\ref{fig1}. The prompt we use for the model is provided in Figure ~\ref{fig2}.

%There are four classes which are as follows misrepresentation, fabrication, false\_attribution, and incorrect\_quantities\\

%We use the following prompt for the purpose of error classification:\\

%\noindent\fbox{%
%    \parbox{1.0\textwidth}{%
%\begin{minipage}{40em}
    
%The task is centered around identifying factual errors in machine-generated summaries. \\

%You will be provided with an article within \{\} \\

%The corresponding machine-generated summary within <>\\

%The objective is to classify every data point.\\

%<Definition for the error class>\\

%The article is-\{Main Document\}\\

%The machine-generated summary is-<incorrect summary>\\

%Please suggest Yes if the machine-generated %summary belongs to XX class or suggest No\\
%\end{minipage}
%}
%}

%}%
%}

%Here, XX can be misrepresentation, fabrication, false\_attribution, and incorrect\_quantities

\subsection{Algorithmic approach}
Here we discuss the various prompt based algorithmic approaches that we took to attempt the problem of multi-label error classification:-
\footnote{We want to specify that all our results are non-deterministic in nature i.e. the same hyperparameter settings can lead to different results on different test runs.}
\begin{itemize}
\item[--] \textbf{\textit{Algorithm 1}} tries to prompt the LLM to understand whether the given incorrect summary belongs to the class misrepresentation. Then the labels fabrication, false\_attribution, and incorrect\_quantities are checked, respectively. If we get one predicted label for a given incorrect summary, we stop the algorithm. The simple heuristic behind first checking whether the (incorrect summary, original document) pair belongs to class misrepresentation is the fact that the misrepresentation class occurs in higher proportions in the training data.
The pseudocode of the algorithm is given in Algorithm~\ref{algorithm1}.
\item[--] \textbf{\textit{Algorithm 2}} tries to prompt the LLM in order to understand whether the given incorrect summary belongs to the class false\_attribution. Then the labels misrepresentation, fabrication, and incorrect\_quantities are checked, respectively. If we get one predicted label for a given incorrect summary, we stop the algorithm. The simple heuristic behind first checking whether the (incorrect summary, original document) pair belongs to the false\_attribution class is the fact that the false\_attribution class occurs in higher proportions in the training data. The pseudocode of the algorithm is given in Algorithm~\ref{algorithm2}.
\item[--] \textbf{\textit{Algorithm 3}} tries to prompt the LLM in order to understand whether the given incorrect summary belongs to the class misrepresentation. Then, the labels false\_attribution, fabrication, and incorrect\_quantities are checked, respectively. If we get two predicted labels for a given incorrect summary, we stop the algorithm. The simple heuristic behind first checking whether the (incorrect summary, original document) pair belongs to the misrepresentation, and false\_attribution class is the fact that the misrepresentation, and false\_attribution class occurs in higher proportions in the training data. The pseudocode of the algorithm is given in 
Algorithm~\ref{algorithm3}. 
\item[--] \textbf{\textit{Algorithm 4}} tries to prompt the LLM in order to understand whether the given incorrect summary belongs to the class misrepresentation. Then the labels fabrication, false\_attribution, and incorrect\_quantities are checked, respectively. If we get two predicted labels for a given incorrect summary we stop the algorithm.The simple heuristic behind first checking whether the (incorrect summary, original document) pair belongs to the misrepresentation, and fabrication class is the fact that the misrepresentation, and fabrication class occurs in higher proportions in the training data.  The pseudocode of the algorithm is given in Algorithm~\ref{algorithm4}. 
\item[--] \textbf{\textit{Algorithm 5}} tries to prompt the LLM in order to understand whether the given incorrect summary belongs to the class false\_attribution. Then the labels misrepresentation, fabrication, and incorrect\_quantities are checked, respectively. If we get four predicted labels for a given incorrect summary, we stop the algorithm. There can be data points for which we get less than four correct data points. We run GPT-3.5 Turbo at different temperatures 0.5, 0.6, 0.7, 0.8, and 0.9 respectively. Then we take an ensemble of the five output test runs being run at different temperatures by considering all the labels that occurred at least twice for a particular datapoint. The ensembling method that we tried helped in providing improved accuracy and generalization bringing more stability into the nature of outputs.
The pseudocode of the algorithm is given in Algorithm~\ref{algorithm5}.
\end{itemize}

\section{Results}

Table~\ref{tab:comparative-metrics-rouge} shows the macro-F1 score considering both correct and incorrect labels.
Table~\ref{tab:comparative-metrics-rouge2} shows the macro-F1 score considering only correct labels. We tried five different prompting-based algorithms. The best result is obtained for Algorithm 5 (Ensembling approach) which we explored in both Table~\ref{tab:comparative-metrics-rouge} and Table~\ref{tab:comparative-metrics-rouge2}.

\begin{table*}[h]
\small
\centering
\scalebox{0.9} {
\begin{tabular}{l|l|l|l}
\toprule
\textbf{Team Name} & \textbf{Method} & \textbf{Run No.} & \textbf{Macro-F1} 
\\
\midrule

%\multicolumn{2}{|c|}{\textbf{General domain LLMs}} 
%\\ \hline

Text Titans & Algorithm 1 & 1 & 0.044 \\ \midrule

Text Titans & Algorithm 2 & 2 & 0.024\\ \midrule

Text Titans & Algorithm 3 & 3 & 0.089\\ \midrule

Text Titans & Algorithm 4 & 4 & 0.112\\ \midrule

Text Titans & Algorithm 5(Ensemble) & 5 & \textbf{0.156}\\ \bottomrule

%Algorithm 1  & \\ \hline
\end{tabular} }

%}
\caption{Macro-F1 score considering both correct and incorrect labels.}
%Entries with an asterisk (*) indicate a value that is statistically significantly higher in terms of the student T-test at a 95\% confidence interval than the best value achieved by an extractive summarization model.}
\label{tab:comparative-metrics-rouge}
\end{table*}

\begin{table*}[h]
\small
\centering
\scalebox{0.9}{
\begin{tabular}{l|l|l|l}
\toprule
\textbf{Team Name} & \textbf{Method} & \textbf{Run No.} &\textbf{Macro-F1} 
\\
\midrule

%\multicolumn{2}{|c|}{\textbf{General domain LLMs}} 
%\\ \hline

Text Titans & Algorithm 1 & 1 & 0.152\\ \midrule

Text Titans & Algorithm 2 & 2 & 0.093\\ \midrule

Text Titans & Algorithm 3 & 3 & 0.291\\ \midrule

Text Titans & Algorithm 4 & 4 & 0.355\\ \midrule

Text Titans & Algorithm 5(Ensemble) & 5 & \textbf{0.527}\\ \bottomrule

%Algorithm 1  & \\ \hline
\end{tabular}}

%}
\caption{Macro-F1 score considering only correct labels.}
%Entries with an asterisk (*) indicate a value that is statistically significantly higher in terms of the student T-test at a 95\% confidence interval than the best value achieved by an extractive summarization model.}
\label{tab:comparative-metrics-rouge2}
\end{table*}

\begin{algorithm}
\caption{Pseudocode}
\label{algorithm1}
$counter(variables)\gets 0 $\;
For {all data points:}\\
\eIf{prompted LLM to check whether the given article and the corresponding summary belong to class misrepresentation}
{
    Then output the class misrepresentation
	counter(variable) = counter(variable)+1

 \eIf{counter(variable)==1 for a datapoint}
{
    stop checking for the next label for that datapoint
}
{
Do not perform any action
}
}
{
Do not perform any action
}
For {all data points:}\\
\eIf{prompted LLM to check whether the given article and the corresponding summary belong to class fabrication}
{
    Then output the class fabrication
	counter(variable) = counter(variable)+1

 \eIf{counter(variable)==1 for a datapoint}
{
    stop checking for the next label for that datapoint
}
{
Do not perform any action
}
}
{
Do not perform any action
}
For {all data points:}\\
\eIf{prompted LLM to check whether the given article and the corresponding summary belong to class false\_attribution}
{
    Then output the class false\_attribution
	counter(variable) = counter(variable)+1

 \eIf{counter(variable)==1 for a datapoint}
{
    stop checking for the next label for that datapoint
}
{
Do not perform any action
}
}
{
Do not perform any action
}
For {all data points:}\\
\eIf{prompted LLM to check whether the given article and the corresponding summary belong to class incorrect\_quantities}
{
    Then output the class incorrect\_quantities
	counter(variable) = counter(variable)+1

 \eIf{counter(variable)==1 for a datapoint}
{
    stop checking for the next label for that datapoint
}
{
Do not perform any action
}
}
{
Do not perform any action
}
\end{algorithm}

\begin{algorithm}
\caption{Pseudocode}
\label{algorithm2}
$counter(variables)\gets 0 $\;
For {all data points:}\\
\eIf{prompted LLM to check whether the given article and the corresponding summary belong to class false\_attribution}
{
    Then output the class false\_attribution
	counter(variable) = counter(variable)+1

 \eIf{counter(variable)==1 for a datapoint}
{
    stop checking for the next label for that datapoint
}
{
Do not perform any action
}
}
{
Do not perform any action
}
For {all data points:}\\
\eIf{prompted LLM to check whether the given article and the corresponding summary belong to class misrepresentation}
{
    Then output the class misrepresentation
	counter(variable) = counter(variable)+1

 \eIf{counter(variable)==1 for a datapoint}
{
    stop checking for the next label for that datapoint
}
{
Do not perform any action
}
}
{
Do not perform any action
}
For {all data points:}\\
\eIf{prompted LLM to check whether the given article and the corresponding summary belong to class fabrication}
{
    Then output the class fabrication
	counter(variable) = counter(variable)+1

 \eIf{counter(variable)==1 for a datapoint}
{
    stop checking for the next label for that datapoint
}
{
Do not perform any action
}
}
{
Do not perform any action
}
For {all data points:}\\
\eIf{prompted LLM to check whether the given article and the corresponding summary belong to class incorrect\_quantities}
{
    Then output the class incorrect\_quantities
	counter(variable) = counter(variable)+1

 \eIf{counter(variable)==1 for a datapoint}
{
    stop checking for the next label for that datapoint
}
{
Do not perform any action
}
}
{
Do not perform any action
}
\end{algorithm}

\begin{algorithm}
\caption{Pseudocode}
\label{algorithm3}
$counter(variables)\gets 0 $\;
For {all data points:}\\
\eIf{prompted LLM to check whether the given article and the corresponding summary belong to class misrepresentation}
{
    Then output the class misrepresentation
	counter(variable) = counter(variable)+1

 \eIf{counter(variable)==2 for a datapoint}
{
    stop checking for the next label for that datapoint
}
{
Do not perform any action
}
}
{
Do not perform any action
}
For {all data points:}\\
\eIf{prompted LLM to check whether the given article and the correponding summary belongs to class false\_attribution}
{
    Then output the class false\_attribution
	counter(variable) = counter(variable)+1

 \eIf{counter(variable)==2 for a datapoint}
{
    stop checking for the next label for that datapoint
}
{
Do not perform any action
}
}
{
Do not perform any action
}
For {all data points:}\\
\eIf{prompted LLM to check whether the given article and the corresponding summary belong to class fabrication}
{
    Then output the class fabrication
	counter(variable) = counter(variable)+1

 \eIf{counter(variable)==2 for a datapoint}
{
    stop checking for the next label for that datapoint
}
{
Do not perform any action
}
}
{
Do not perform any action
}
For {all data points:}\\
\eIf{prompted LLM to check whether the given article and the corresponding summary belong to class incorrect\_quantities}
{
    Then output the class incorrect\_quantities
	counter(variable) = counter(variable)+1

 \eIf{counter(variable)==2 for a datapoint}
{
    stop checking for the next label for that datapoint
}
{
Do not perform any action
}
}
{
Do not perform any action
}
\end{algorithm}

\begin{algorithm}
\caption{Pseudocode}
\label{algorithm4}
$counter(variables)\gets 0 $\;
For {all data points:}\\
\eIf{prompted LLM to check whether the given article and the corresponding summary belong to class misrepresentation}
{
    Then output the class misrepresentation
	counter(variable) = counter(variable)+1

 \eIf{counter(variable)==2 for a datapoint}
{
    stop checking for the next label for that datapoint
}
{
Do not perform any action
}
}
{
Do not perform any action
}
For {all data points:}\\
\eIf{prompted LLM to check whether the given article and the corresponding summary belong to class fabrication}
{
    Then output the class fabrication
	counter(variable) = counter(variable)+1

 \eIf{counter(variable)==2 for a datapoint}
{
    stop checking for the next label for that datapoint
}
{
Do not perform any action
}
}
{
Do not perform any action
}
For {all data points:}\\
\eIf{prompted LLM to check whether the given article and the corresponding summary belong to class false\_attribution}
{
    Then output the class false\_attribution
	counter(variable) = counter(variable)+1

 \eIf{counter(variable)==2 for a datapoint}
{
    stop checking for the next label for that datapoint
}
{
Do not perform any action
}
}
{
Do not perform any action
}
For {all data points:}\\
\eIf{prompted LLM to check whether the given article and the correponding summary belongs to class incorrect\_quantities}
{
    Then output the class incorrect\_quantities
	counter(variable) = counter(variable)+1

 \eIf{counter(variable)==2 for a datapoint}
{
    stop checking for the next label for that datapoint
}
{
Do not perform any action
}
}
{
Do not perform any action
}
\end{algorithm}

\begin{algorithm}
\caption{Pseudocode}
\label{algorithm5}
$counter(variables)\gets 0 $\;
For {all data points:}\\
\eIf{prompted LLM to check whether the given article and the corresponding summary belong to class misrepresentation}
{
    Then output the class misrepresentation
	counter(variable) = counter(variable)+1

 \eIf{counter(variable)==4 for a datapoint}
{
    stop checking for the next label for that datapoint
}
{
Do not perform any action
}
}
{
Do not perform any action
}
For {all data points:}\\
\eIf{prompted LLM to check whether the given article and the corresponding summary belong to class false\_attribution}
{
    Then output the class false\_attribution
	counter(variable) = counter(variable)+1

 \eIf{counter(variable)==4 for a datapoint}
{
    stop checking for the next label for that datapoint
}
{
Do not perform any action
}
}
{
Do not perform any action
}
For {all data points:}\\
\eIf{prompted LLM to check whether the given article and the corresponding summary belong to class fabrication}
{
    Then output the class fabrication
	counter(variable) = counter(variable)+1

 \eIf{counter(variable)==4 for a datapoint}
{
    stop checking for the next label for that datapoint
}
{
Do not perform any action
}
}
{
Do not perform any action
}
For {all data points:}\\
\eIf{prompted LLM to check whether the given article and the corresponding summary belong to class incorrect\_quantities}
{
    Then output the class incorrect\_quantities
	counter(variable) = counter(variable)+1

 \eIf{counter(variable)==4 for a datapoint}
{
    stop checking for the next label for that datapoint
}
{
Do not perform any action
}
}
{
Do not perform any action
}
\end{algorithm}

\clearpage
\section{Conclusion and Future Work}
We were given the task of multi-label error classification where we had to classify a document into four classes namely misrepresentation, fabrication, false\_attribution, and incorrect\_quantities. We tried several prompt-based algorithmic approaches for the multi-label error classification task that we were given as a part of Task 2. We observed that the Algorithm 5 (Ensembling approach) that we explored obtained the best results. Future work would involve trying a few-shot technique and trying larger language models such as GPT-4.
\bibliography{sample-ceur}
\end{document}